%% file: main.tex
\documentclass[11pt,a4paper]{article}
\usepackage[hyperref]{eacl2021}
\usepackage{times}
\usepackage{latexsym}

\usepackage{microtype}

\aclfinalcopy %
\def\aclpaperid{} %

\usepackage[utf8]{inputenc}
\usepackage{enumitem}
\usepackage{subfiles}
\usepackage{graphicx}
\usepackage{caption}
\usepackage{subcaption}
\usepackage{booktabs}
\usepackage{float}
\usepackage{amssymb}
\usepackage{mathtools}
\usepackage{amsmath}
\usepackage{multirow}
\usepackage{tabu}
\usepackage{pifont}
\usepackage{longtable}
\usepackage{xspace}

\usepackage[capitalise,noabbrev]{cleveref}
\usepackage{todonotes}
\usepackage[english]{babel}
\usepackage[autostyle]{csquotes}
\MakeOuterQuote{"}

\graphicspath{{./figures/}}
\newcommand{\sota}{state-of-the-art\xspace}
\newcommand{\eg}{\emph{e.g.}\xspace}
\newcommand{\ie}{\emph{i.e.}\xspace}

\newcommand{\syelp}{\emph{Syelp}\xspace}
\newcommand{\fyelp}{\emph{Fyelp}\xspace}
\newcommand{\samazon}{\emph{SAmazon}\xspace}
\newcommand{\rt}{\emph{RottenTomatoes}\xspace}

\newcommand{\xhat}{$\hat x$\xspace}

\newcommand{\yhat}{$\hat y$\xspace}
\newcommand{\xstar}{$x^{*}$\xspace}

\newcommand{\dae}{\textsc{Dae}\xspace}
\newcommand{\bt}{\textsc{Bt}\xspace}
\newcommand{\adv}{\textsc{Adv}\xspace}
\newcommand{\mrt}{\textsc{Mrt}\xspace}

\newcommand{\trans}{Transformer\xspace}
\newcommand{\rnn}{RNN\xspace}

\newcommand{\acc}{\textsc{Acc}\xspace}
\newcommand{\emd}{\textsc{EMD}\xspace} 
\newcommand{\wmd}{\textsc{WMD}\xspace}
\newcommand{\wms}{\textsc{WMS}\xspace}
\newcommand{\bleu}{\textsc{BLEU}\xspace}
\newcommand{\sbleu}{\textsc{sBLEU}\xspace}
\newcommand{\ppl}{\textsc{PPL}\xspace}

\title{Empirical Evaluation of Supervision \\ Signals for Style Transfer Models} \author{
  Yevgeniy Puzikov$^1$,\ Stanley Simoes\thanks{~~Work done during an internship at UKP Lab.},\ Iryna Gurevych$^1$,\ Immanuel Schweizer$^2$ \\
  $^1$ Ubiquitous Knowledge Processing Lab (UKP Lab), \\
  Department of Computer Science, Technical University of Darmstadt \\
  $^2$ Merck KGaA, Darmstadt, Germany \\
  \url{https://www.ukp.tu-darmstadt.de} \\
  {\tt \small stanleysimoes@gmail.com} \\
  {\tt \small immanuel.schweizer@merckgroup.com} }

\date{}

\begin{document}
\maketitle

\begin{abstract}
  Text style transfer has gained increasing attention from the
  research community over the recent years. However, the proposed
  approaches vary in many ways, which makes it hard to assess the
  individual contribution of the model components. In style transfer,
  the most important component is the optimization technique used to
  guide the learning in the absence of parallel training data. In this
  work we empirically compare the dominant optimization paradigms
  which provide supervision signals during training: backtranslation,
  adversarial training and reinforcement learning. We find that
  backtranslation has model-specific limitations, which inhibits
  training style transfer models. Reinforcement learning shows the
  best performance gains, while adversarial training, despite its
  popularity, does not offer an advantage over the latter alternative.
  In this work we also experiment with Minimum Risk
  Training~\cite{och2003minimum}, a popular technique in the machine
  translation community, which, to our knowledge, has not been
  empirically evaluated in the task of style transfer. We fill this
  research gap and empirically show its efficacy.
\end{abstract}

\section{Introduction}\label{sec:introduction}
Text style transfer is the task of changing stylistic properties of an
input text, while retaining its style-independent
content. Regenerating existing text to cater to a target audience has
diverse use-cases such as rewriting offensive language on social
media~\cite{dossantos2018fighting}; making a text more
formal~\cite{rao2018dear}, romantic~\cite{li2018delete} or
politically-slanted~\cite{prabhumoye2018style}; changing its
tense~\cite{ficler2017controlling} or sentiment~\cite{shen2017style}.

While training unsupervised models for \emph{generating} style-infused
texts can be done using conditional language-modelling techniques, in
order to perform style \emph{transfer}, one needs to find a source of
supervision signal. Parallel corpora for this task are
scarce~\cite{xu2012paraphrasing,jhamtani2018learning,rao2018dear,kang2019male},
so researchers focused on finding non-parallel supervision signals.

We analyzed previous work and came to a conclusion that, although many
approaches have been proposed, they all employ similar optimization
methods that form groups of techniques; one simply combines them to
produce a style-transfer model. The ``recipe'' is using denoising
autoencoding as a mechanism to teach the model to generate grammatical
texts; style-infusion comes from: 1) discriminator-based training; 2)
backtranslation; 3) metric supervision via reinforcement learning
(RL). Our work examines the properties of these methods and finds
which of them contribute to the success or failure of a style-transfer
approach.

Our contributions are three-fold:
\begin{itemize}
  \item We provide a structured overview of the supervision techniques used for training style transfer models.
  \item We find evidence of the limitations of the existing techniques.
  \item To the best of our knowledge, we are the first ones to use Minimum Risk Training
    technique~\cite{och2003minimum} in style transfer. We prove its efficacy in the subsequent experiments.
\end{itemize}

In what follows, we first describe the notation used throughout the
paper, then introduce each of the examined model components. After that,
we explain our experimental setup, analyze the results and pinpoint the approaches'
limitations.

\section{Overview}\label{sec:model-description}

We assume that our training data consists of
text--style pairs ($x,s$), where $x$ is a text and $s=(s_1,\dots,s_m)$ is a set of
style values which $x$ has. Each $s_k$ is a discrete value in the set
$\mathcal{S}_k$ of possible values for attribute $k$.

Our task is to learn a mapping from a pair of an input text $x$ and
arbitrary style $\hat s$, to a new text $\hat x$ that exhibits styles
$\hat s$, but has the content of $x$.  Research literature does not
define precisely what content is; usually it is assumed that content
is style-independent. However, whether it is possible to decouple the
two is a topic of an ongoing
debate~\cite{lample2019multiple,john2019disentangled}. In this work,
content is defined as anything in $x$ which does not depend on the
style attributes.

All works we have examined employ some variant of a recurrent neural
network (\rnn, \newcite{rumelhart1986learning}) or
\trans~\cite{vaswani2017attention} as a text generator. For
simplicity, as a generator network, we implemented a bi-directional
encoder and uni-directional decoder with Gated Recurrent
Unit~\cite{cho2014gru}, attention~\cite{bahdanau2014attn}, and the
pooling mechanism of \newcite{lample2019multiple}. The generator model
first encodes text $x$ into a latent representation $z = e(x)$, then
decodes $(z, \hat s)$ into $\hat x = d(z, \hat s)$, where $e$ and $d$
are encoder and decoder parts of the model.

What differs between the approaches which we compare in this paper is
the optimization technique used to train the model. These techniques
are described in the following subsections. Hyperparameter values are
reported in \cref{sec:supp-training-details}.

\subsection{Autoencoding}\label{sec:autoencoding}
\begin{figure}[h]
  \centering
  \includegraphics[width=0.25\textwidth]{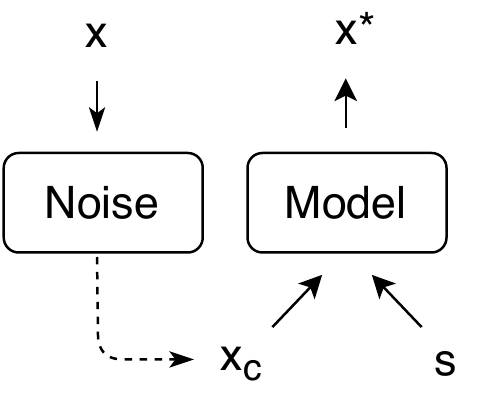}
  \caption{Schematic view of the \dae training procedure. $x$ is the
    input text, $x_c$ is its noised version, \xstar is the
    reconstruction of $x$.  The dashed line shows the absence of any
    transformation, \ie, that the output of the noising procedure
    becomes the input to the model at the next step.}
  \label{fig:ae}
\end{figure}

First, the model is trained with a denoising autoencoding (\dae)
objective to learn to produce grammatical texts from corrupted
inputs. An illustration of this process is shown in
\Cref{fig:ae}. Following~\newcite{lample2019multiple}, we corrupt a
given text $x$ by randomly dropping and shuffling words, which
produces $x_c$. The corrupted text serves as input to the encoder; the
target sequence to reconstruct is the original text $x$. \dae training
minimizes the following objective:
\begin{equation}
  L_{ae} = -\log P \Big( x | e(x_c), s \Big)
\end{equation}

\subsection{Backtranslation}\label{sec:backtranslation}
Backtranslation (\bt) was originally proposed
by~\newcite{sennrich2016improving} in the context of machine
translation as a method of creating silver-standard data and
bootstrapping machine translation models. Some researchers
successfully applied it to style transfer, but used it in different
ways. \newcite{zhang2020parallel} employed \bt to obtain additional
training data, while \newcite{lample2019multiple} treated it as a
source of indirect supervision, arguing that \bt helps to prevent the
model from doing just reconstruction.

Interestingly enough, \newcite{prabhumoye2018style} used \bt to do the
opposite. The authors refer to a study of
~\newcite{rabinovich2017personalized} who showed that stylistic
properties are obfuscated by both manual and automatic machine
translation, \ie, backtranslation can be used to rephrase a text while
reducing its stylistic properties. It seems that sometimes \bt
exhibits an additional supervision signal \cite{lample2019multiple},
and sometimes it has a regularization
effect~\cite{rabinovich2017personalized, prabhumoye2018style}.

\begin{figure}[h]
  \centering
  \includegraphics[width=0.35\textwidth]{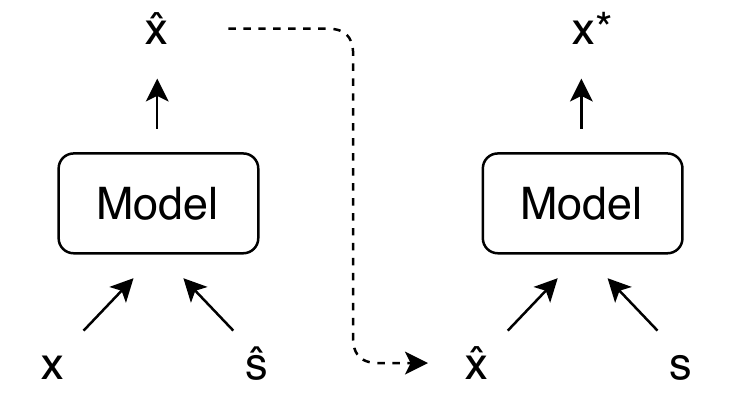}
  \caption{Schematic view of the \bt training procedure.
    $x$ is the input text, 
    $s$ is the corresponding input style, $\hat s$ is the desired style.
    \xhat and \xstar are generated outputs. 
    "Model" is the encoder-decoder generator.}  
  \label{fig:bt}
\end{figure}

An illustration of the backtranslation process for style transfer is
shown in \Cref{fig:bt}.  Given an input text $x$ and original style
$s$, we first perturb $s$ by changing at least one of the attributes
to produce $\hat s$. Next, the model takes ($x$,$\hat s$) as input and
generates text $\hat x$. The model then uses $\hat x$ and the original
style $s$ to produce \xstar which, ideally, is a reconstruction of
$x$. \bt training minimizes the following objective:

\begin{equation}
  L_{bt} = - \log P \bigg( x | e \Big( d \big( e(x), \hat s \big) \Big), s \bigg)
\end{equation}

With backtranslation, the training alternates between an autoencoding
and backtranslation steps. The final optimization function we minimize
is a linear combination of \dae and \bt losses:

\begin{equation}\label{eq:1}
  L_{total} = \lambda_{ae} L_{ae} + \lambda_{bt} L_{bt}
\end{equation}

The $\lambda$ parameters constitute a trade-off between performing
more content preservation or style transfer. In our experiments we
follow~\cite{lample2019multiple} and anneal the $\lambda_{ae}$ to 0
towards the end of training, while keeping $\lambda_{bt}$ equal to 1.

\subsection{Adversarial Training}\label{sec:adversarial-training}

Adversarial training~\cite{goodfellow2014gan} provides means for
leveraging training signals from non-parallel corpora for style
transfer. One popular approach in this direction is to disentangle the
input text's content and style information by employing adversarial
networks that operate on the input text's latent representation, \ie,
the encoder output. This can be done by separating the latent
representation into the content representation and style
representation~\cite{john2019disentangled}, or learning
style-invariant latent representations~\cite{fu2018style}. Another
approach is to use an adversarial network within the backtranslation
framework~\cite{logeswaran2018content,dai2019styletransformer}, which
is what we employed in our experiments. Using adversarial
discriminators in such a scenario helps matching the distribution of
style-specific latent representations of real vs. synthetic
texts~\cite{shen2017style}.

\begin{figure}[h]
  \centering
  \includegraphics[width=0.35\textwidth]{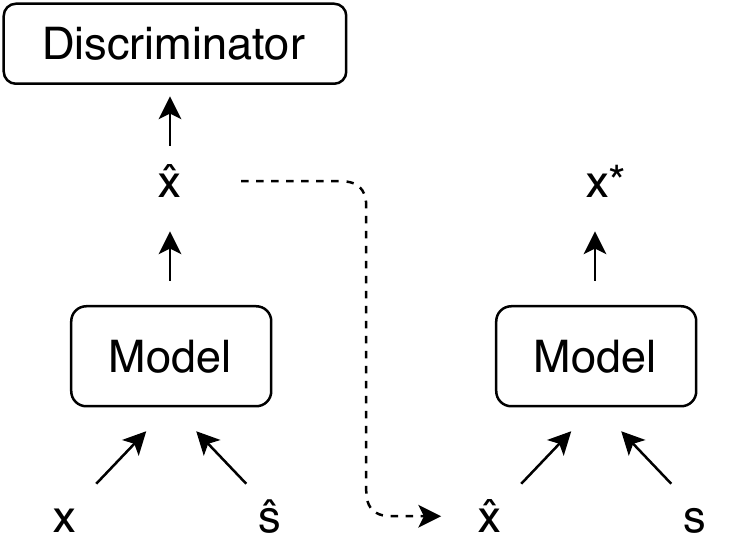}
  \caption{Schematic view of the \adv training procedure.
    The inputs and outputs are the same as for the \bt stage.}  
  \label{fig:adv}
\end{figure}

An illustration of the adversarial training of the generator model is
shown in \Cref{fig:adv}. We implement the multi-class discriminator
of~\newcite{dai2019styletransformer} using a GRU-based encoder with a
classification layer which predicts the style of $\hat x$. Adversarial
training involves alternating between training the generator model to
produce style-infused texts, and training the discriminator to
distinguish between real sentences of different styles, on one hand,
and model-generated texts, on the other hand. Training the latter is
straightforward; we follow~\newcite{dai2019styletransformer} and refer
the reader to the original paper for details.

When training both the discriminator and the generator, we minimize
the cross entropy loss, and teach the discriminator to predict style,
given a text (either real one or generated by the model), and the
generator to output texts that look \emph{real}, \ie, similar to the
texts with the desired style in the training data.

Note that adding the adversarial component is done on top of \bt
model, because with \dae and \adv only, it is not possible to force
the model to preserve the content. For this reason, training the
generator now consists of three terms:
\begin{align*}
  L_{adv} &= - \log P_D (\hat s | \hat x) \\
  L_{total} &= \lambda_{ae} L_{ae} + \lambda_{bt} L_{bt} + \lambda_{adv} L_{adv}
\end{align*}

We reuse the same $\lambda$ parameters as in the \bt approach. $\lambda_{adv}$ is set to 1.0.~\footnote{This seemed to be a reasonable value; we did not perform any hyperparameter tuning.}

\begin{table*}[t]
  \centering
  \small
  \begin{tabular}{lcc|cc|ccccc}
    \toprule        
    & \multicolumn{2}{c|}{Sentiment} & \multicolumn{2}{c|}{Gender} & \multicolumn{5}{c}{Category}                       \\
    \midrule                     
    \textbf{FYelp}             & Positive   & Negative   & Male       & Female     & American   & Asian      & Bar        & Dessert   & Mexican  \\
    & 1,035,609  & 197,203    & 584,637    & 648,175    & 338,899    & 208,483    & 372,873    & 209,949   & 102,608  \\

    \textbf{RottenTomatoes}    & Positive   & Negative   & Male       & Female     & Critic      & Audience &- &- &- \\
    & 245,241    & 118,857    & 268,564    & 95,535     & 77,467      & 286,631  &- &- &- \\
    
    \midrule
    \textbf{SYelp}               & Positive   & Negative   &- &- &- &- &- &- &- \\
    & 266,041    & 177,218    &- &- &- &- &- &- &- \\
    
    \textbf{SAmazon}             & Positive   & Negative   &- &- &- &- &- &- &- \\
    & 277,228    & 277,769    &- &- &- &- &- &- &- \\         
    \bottomrule

  \end{tabular}
  \caption{The number of training instances per attribute for each
    dataset. Preprocessing details are given in the appendix.}
\label{tab:datasets}
\end{table*}

\subsection{Minimum Risk Training}\label{sec:reinf-learn}
Existing works have also explored architectures based on RL techniques
for text style transfer. For example, \newcite{gong2019reinforcement}
use evaluation metrics for style, content preservation, and
naturalness as the training objective within the RL
framework. \newcite{wu2019hierarchical} use a hierarchical model where
the high-level agent decides where the input text needs to be
modified, and the low-level agent decides on the modification.

Following the success of the Minimum Risk Training
method~\cite{och2003minimum} in the machine translation community, we
decided to experiment with it as a potential candidate of the RL
techniques. Since the advent of neural networks, there have been
successful attempts to use \mrt for generation tasks, like neural
machine translation~\cite{gao2014learning,shen2016minimum}, but we are
unaware of any work that has explored its utility in the domain of
style transfer. Yet, it has a number of advantages over other RL
alternatives. First, it is very easy to implement and use. Second,
unlike other RL algorithms, like
\textsc{REINFORCE}~\cite{williams1992reinforce}, \mrt uses multiple
examples at a time to estimate risk. This allows for efficient data
batching, leading to faster training speed and diversity in the
generated examples.

\begin{figure}[h]
  \centering
  \includegraphics[width=0.45\textwidth]{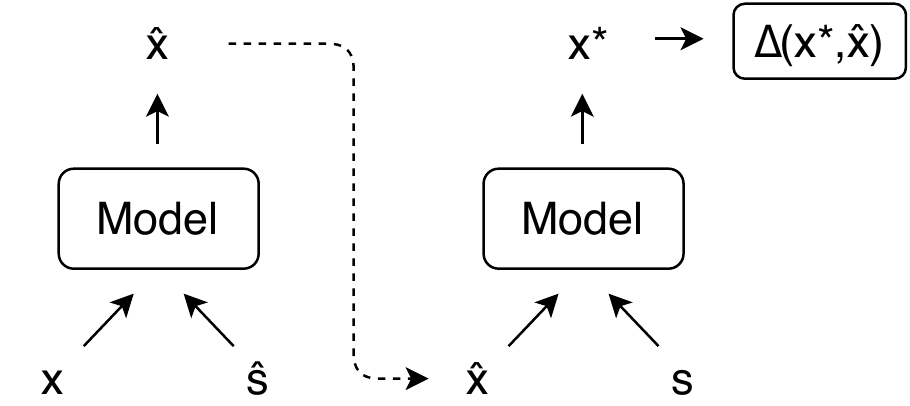}
  \caption{Schematic view of the \mrt training procedure.
    The inputs and outputs are the same as for the \bt stage.}  
  \label{fig:mrt}
\end{figure}

An illustration of the \mrt training step is shown in
\Cref{fig:mrt}. Note that it is performed on top of the \bt procedure,
since we want the outputs to be similar in content with the input
text. The main idea is to use evaluation metrics (possibly
non-differentiable) as loss functions and assume that the optimal set
of model parameters should minimize the expected loss on the training
data. Given an input $x$, a model prediction \yhat, a desired output
$y$ and a loss function $\Delta(\hat y,y)$, \mrt seeks a posterior
$P(\hat {y}|x)$ to minimize the expected loss
$\mathbf{E}_{\hat {y} \sim P(\hat {y}|x)} \Delta (\hat {y},y)$.

Since we do not have reference outputs, we cannot use reference-based
metrics. However, we can use style intensity classifiers to compute a
metric that could guide the model towards generating better
outputs. According to \newcite{mir2019evaluating}, when evaluating
style intensity, the metric that correlates most with human
judgements, is direction-corrected Earth-Mover's Distance
(\emd)~\cite{rubner1998ametric}. We measure it between the style
distributions of the texts generated during the backtranslation
process (see \Cref{sec:evaluation} for details):
\begin{align*}
    L_{mrt}   &= E_{x^{*} \sim P(x^{*} | \hat {x})} \Delta (x^{*},\hat {x}) \\
    L_{total} &= \lambda_{ae} L_{ae} + \lambda_{bt} L_{bt} + \lambda_{mrt} L_{mrt}
\end{align*}

We use the same $\lambda$ hyperparameters as in the \bt and \adv
cases, $\lambda_{mrt}$ is set to 1.0.

\section{Experimental Setup}\label{sec:experiments}
\subsection{Datasets}\label{sec:datasets}

Following previous work, we used publicly available Yelp restaurant
and Amazon product review datasets which vary in one attribute, the
review sentiment~\cite{shen2017style,li2018delete}.

We followed \newcite{lample2019multiple} and included a
multi-attribute version of the Yelp restaurant review dataset which
contains texts varying in product categories, gender of the reviewers,
and sentiment of the review. We also added a multi-attribute dataset
of \newcite{ficler2017controlling} which contains movie reviews from
the Rotten Tomatoes website. The texts vary in professionality and
sentiment dimensions. We also added gender annotations, following the
same procedure as for the \fyelp dataset.

The lengths of \rt and \fyelp texts vary a lot --- some exceed 1k
tokens. Due to computational limitations, we had to restrict ourselves
to texts no longer than 50 tokens for both datasets; \syelp and
\samazon datasets were not trimmed in any way. The number of training
instances per category for each of the four datasets are shown in
\Cref{tab:datasets}. The details of the preprocessing steps for all
datasets are given in \Cref{sec:supp-dataset-preparation}.

\begin{table*}[t!]
  \centering
  \small
  \begin{tabular}{l|cccccc}
    \toprule
    Model                                       &  \acc(\%) & \emd   & \bleu & \sbleu & \wms & \ppl   \\
    \midrule                                                                           
    CrossAligned \cite{shen2017style}           &  73.8  & 0.68  & 3.3  & 13.2  & 0.66 & 69.1  \\
    Style Embedding \cite{fu2018style}          &  9.1   & 0.05  & \textbf{12.1} & \textbf{69.2}  & \textbf{0.86} & 76.0  \\
    MultiDecoder \cite{fu2018style}             &  46.5  & 0.42  & 7.4  & 37.8  & 0.72 & 146.7     \\
    TemplateBased \cite{li2018delete}           &  81.1  & 0.74  & 11.1 & 44.2  & 0.70 & 1915.0     \\
    RetrieveOnly \cite{li2018delete}            &  \textbf{93.8}  & \textbf{0.84}  & 0.4  & 0.7   & 0.52 & \textbf{7.9}       \\
    DeleteOnly \cite{li2018delete}              &  83.5  & 0.76  & 7.6  & 28.6  & 0.68 & 71.5        \\
    DeleteAndRetrieve \cite{li2018delete}       &  87.2  & 0.79  & 8.5  & 29.1  & 0.67 & 86.0       \\
    \midrule                                                                           
    \dae                                        &  24.5  & 0.20  & 11.7 & 58.1  & 0.86 & 51.8  \\
    \dae $+$ \bt                                &  85.8  & 0.79  & 6.8  & 21.4  & 0.70 & 42.8  \\
    \dae $+$ \bt $+$ \adv                       &  87.2  & 0.80  & 6.9  & 20.7  & 0.70 & 40.6  \\
    \dae $+$ \bt $+$ \mrt                       &  88.1  & 0.81  & 6.9  & 20.1  & 0.70 & 41.0  \\
    \midrule                                                                           
    Input copy                                  &   3.9  & 0.00 & 18.4  & 100.0 & 1.00 &  8.2      \\
    \bottomrule
  \end{tabular}
  \caption{Automatic metric evaluation results on the \syelp test set
    (lower-cased).  \bleu scores are computed between the test set
    human references and model outputs.  For all scores, except for
    perplexity (PPL): the higher the better.  ACC, BLEU and sBLEU
    values are in range $[0,100]$; EMD in $[0,1]$; WMS in $(0,1]$; PPL
    in $[0, \infty ]$.}
  \label{tab:eval_syelp}
\end{table*}

\subsection{Evaluation Metrics}\label{sec:evaluation}

A lot of work has been done in order to make evaluation of style
transfer models more
reliable~\cite{shen2017style,fu2018style,zhao2018adversarily,li2018delete,mir2019evaluating}. We
combine the evaluation setups of \newcite{lample2019multiple} and
\newcite{mir2019evaluating} in order to make our results comparable to
the previous work; the details are in \Cref{sec:supp-evaluation}. We
evaluate the system outputs across three quality dimensions.

\textbf{Attribute control} is assessed by in-domain
fasttext~\cite{joulin2016bag} classifiers. For each dataset, we use
the train portion of the data to train attribute-specific
classifiers. Given a predicted text, a classifier outputs a
probability distribution over possible styles. We use the
highest-scoring class as a classifier prediction and compare it with
the gold-standard label to compute the accuracy. We also compute \emd
between the probability distributions of the predicted text, on one
hand, and the original text, on the other. Finally, all scores are
averaged across attributes.

\textbf{Fluency} is approximated by the perplexity computed by a
5-gram KenLM model~\cite{heafield2011kenlm} with Kneser--Ney
smoothing~\cite{NEY19941}.

\textbf{Content preservation} is measured by two groups of
metrics. First, we use an embedding-based Word-Mover's Similarity
(\wms), the normalized inverse of the Word-Mover's Distance. This is
done in order to make it easier for the reader to compare approaches:
the higher the score, the better (similar to the other metrics). The
second group includes \bleu~\cite{papineni2002bleu} and self-\bleu (or
\sbleu). The \syelp and \samazon test sets have human references, so
we compute \bleu scores between these references and the model
outputs. \fyelp and \rt do not have human references, and we compute
\sbleu scores between the input texts and the generated outputs.

\section{Results}\label{sec:results}

\begin{table*}[t!]
  \centering
  \small
  \begin{tabular}{l|cccccc}
    \toprule
    Model                                       &  \acc(\%) & \emd   & \bleu & \sbleu & \wms & \ppl   \\
    \midrule                                                                           
    CrossAligned \cite{shen2017style}           &  \textbf{74.5}  & \textbf{0.45}  & 0.4  & 0.5   & 0.55 & 20.5  \\ %
    Style Embedding \cite{fu2018style}          &  39.7  & 0.19  & 10.2 & 29.5  & 0.67 & 81.1  \\ %
    MultiDecoder \cite{fu2018style}             &  72.1  & 0.41  & 4.9  & 14.4  & 0.61 & 78.9     \\ %
    TemplateBased \cite{li2018delete}           &  69.9  & 0.40  & 26.6 & 64.0  & 0.78 & 91.1     \\ %
    RetrieveOnly \cite{li2018delete}            &  73.5  & 0.43  & 0.9  & 2.1   & 0.54 & \textbf{7.7}       \\ %
    DeleteOnly \cite{li2018delete}              &  51.0  & 0.26  & 25.4 & 60.9  & 0.80 & 37.7        \\ %
    DeleteAndRetrieve \cite{li2018delete}       &  56.4  & 0.30  & 23.3 & 54.3  & 0.77 & 57.4       \\ %
    \midrule                                                                           
    \dae                                        &  20.2  & 0.03  & 30.2 & \textbf{79.6}  & \textbf{0.94} & 30.1  \\
    \dae $+$ \bt                                &  34.4  & 0.13  & \textbf{30.9} & 78.9  & 0.92 & 29.3  \\
    \dae $+$ \bt $+$ \adv                       &  47.3  & 0.23  & 28.5 & 72.0  & 0.89 & 36.9  \\
    \dae $+$ \bt $+$ \mrt                       &  50.4  & 0.25  & 28.1 & 70.9  & 0.88 & 38.3  \\
    \midrule                                                                           
    Input copy                                  &  17.1  & 0.00  & 38.4 & 100.0 & 1.00 &  8.5      \\
    \bottomrule
  \end{tabular}
  \caption{Automatic metric evaluation results on the \samazon test
    set (lower-cased).  \bleu scores are computed between the test set
    human references and model outputs.}
  \label{tab:eval_samazon}
\end{table*}

\subsection{Single-Attribute (\syelp, \samazon)}\label{sec:single-attr-syelp}

We first evaluate the described methods in the single-attribute
scenario. \Cref{tab:eval_syelp} and \Cref{tab:eval_samazon} show their
performance on the test portion of the \syelp and \samazon datasets,
respectively. The results for previous work are computed based on the
outputs
from~\cite{li2018delete}.\footnote{\url{https://github.com/lijuncen/Sentiment-and-Style-Transfer}.}

The first striking observation is that all models achieve low \bleu
scores. Taking into consideration the high \wms scores of some models,
this suggests that using an n-gram overlap between a human reference
and model output is inadequate for style transfer --- the potential
variability of re-generating text in a different style is too high to
be captured by an overlap with one reference text. This observation is
reinforced by the fact that the models with the best transfer
performance (accuracy and \emd) also exhibit lowest \bleu scores. The
fact that \sbleu and \wms have a large gap indicates that computing an
n-gram overlap between the input and system output is also a very
superficial way of measuring content preservation, calling for the
usage of vector-space models, like \wmd.

Interestingly, the performance of the models proposed in the
literature is not consistent across datasets. \samazon has longer and
more diverse sentences than \syelp, which could explain why template-
and retrieval-based approaches underperform, compared to the
data-driven alternatives. However, it is not clear why both the
previously proposed neural models and the approaches we implemented
and experimented with in this paper show such a large gap between the
results on \syelp and \samazon.

It is surprising that \dae by itself can do some amount of style
transfer, even without the additional supervision signal. This most
likely is the consequence of indiscriminate noising of tokens in the
input text and removing of style-bearing words during the noising
step. The work of~\newcite{shen2019latent} offers a plausible
explanation for that: denoising seems to help autoencoders to map
similar texts to similar latent representation and promote sequence
neighborhood preservation.

Among the tested supervision signals, \mrt has a slight
preference. However, in the single-attribute scenario, the best way to
do style transfer seems to be a simple nearest-neighbour approach
(RetrieveOnly): by retrieving a semantically-similar text with the
desired style from the available corpus.

Manual examination of model predictions revealed that none of the
approaches goes further than replacing several style-bearing
words. This happens due to a limited variation in the data. For
example, \syelp texts are at most 15 tokens long, and most reviews
have similar structure, so the models learn to do minimal edits to
perform style transfer. They also fail when it is needed to go beyond
that. For example, all examined approaches failed to change the style
in the following cases and produce almost unchanged input text as
prediction:
\begin{itemize}[noitemsep]
    \item \emph{i just walked out , called the manager to complain}
    \item \emph{she does n't say anything and just walks away}
\end{itemize}

\subsection{Multi-Attribute (\fyelp, \rt)}\label{sec:multi-attr-fyelp}

\Cref{tab:eval_fyelp} and \Cref{tab:eval_rt} show the performance of
the considered approaches on \fyelp, and \rt data, respectively.

\begin{table*}[t!]
\small
  \centering
  \begin{tabular}{l|ccccc}
    \toprule
    Model                                       &  \acc(\%) & \emd  & \sbleu & \wms  & \ppl \\
    \midrule
    \dae                                        &  13.9    & 0.02    & \textbf{38.5}  & \textbf{0.76} & 67.1   \\
    \dae $+$ \bt                                &  32.2    & 0.24    & 22.9  & 0.69 & \textbf{29.6}  \\
    \dae $+$ \bt $+$ \adv                       &  42.4    & 0.33    & 22.1  & 0.68 & 31.8   \\
    \dae $+$ \bt $+$ \mrt                       &  \textbf{46.8}    & \textbf{0.36}    & 21.5  & 0.68 & 33.1   \\
    \bottomrule
  \end{tabular}
  \caption{Automatic metric evaluation results on the \fyelp test set
    (lower-cased).}
  \label{tab:eval_fyelp}
\end{table*}

\begin{table*}[t!]
  \small
  \centering
  \begin{tabular}{l|ccccc}
    \toprule
    Model                                       &  \acc(\%) & \emd  & \sbleu & \wms  & \ppl \\
    \midrule                                                              
    \dae                                        &  35.1    & 0.015   & \textbf{39.9}  & \textbf{0.78}  & \textbf{73.79}   \\
    \dae $+$ \bt                                &  55.5    & 0.18    & 28.5  & 0.69 & 83.1  \\
    \dae $+$ \bt $+$ \adv                       &  57.6    & 0.20    & 28.2  & 0.69 & 83.2   \\
    \dae $+$ \bt $+$ \mrt                       &  \textbf{59.6}    & \textbf{0.22}    & 25.6  & 0.68 & 98.5   \\
    \bottomrule
  \end{tabular}
  \caption{Automatic metric evaluation results on the \rt test set
    (lower-cased).}
  \label{tab:eval_rt}
\end{table*}

The trends from the single-attribute transfer seem to be present here
as well. The \sbleu and \wms scores achieved by the \dae model are the
highest, which is intuitive --- the model learns to reconstruct the
input.

\begin{figure}[h]
  \centering
  \includegraphics[width=0.45\textwidth]{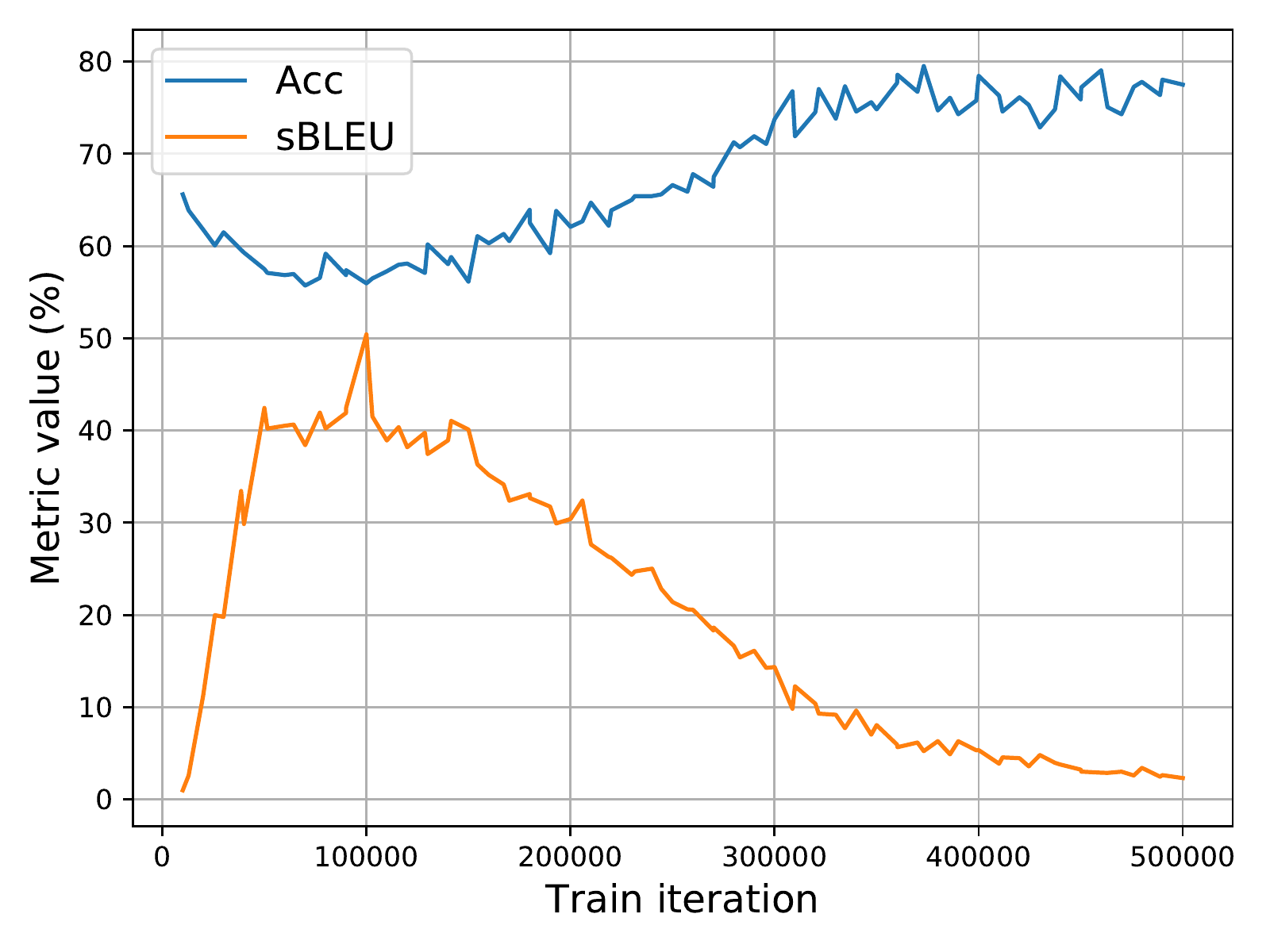}
  \caption{Style transfer accuracy vs \sbleu score during training phase of the \bt model. Data: development set of the \rt dataset.}
  \label{fig:cp-vs-sti}
\end{figure}

The correlation between higher \emd and accuracy scores vs. lower \wmd
and \sbleu scores supports the hypothesis that there is a trade-off
between preserving input content and performing style
transfer. \Cref{fig:cp-vs-sti} shows how content preservation
(measured by \sbleu) and style intensity (\acc) criteria start
competing during \bt model training.

This phenomenon was also observed by \newcite{lai2019multiple}: the
authors note that a model trained longer was better able to transfer
style, but worse at retaining the input’s content.  An evaluation
perspective of this issue was also studied
by~\citet{mir2019evaluating}.  We are not sure whether it is possible
to define a performance upper bound for a particular class of models
and, therefore, deciding which model is \sota in the task of style
transfer is not easy --- the aforementioned trade-off complicates this
issue. This makes finding ways to control this trade-off a very
interesting future research direction.

Quality-wise, \adv and \mrt produce more style-infused instances, but
even they have two flaws. First, they struggle to perform transfer
across all styles simultaneously.  The issue is complicated by the
difficulty of the chosen style attributes themselves. For example,
transferring \emph{gender} style proved to be a challenge even for the
authors of the paper.

The second issue is that models cannot cope with cases when words
usual for one style are used for expressing the opposite style. For
example:
\begin{itemize}[noitemsep]
    \item \emph{there are much better places for breakfast .}
    \item \emph{anything they say , ask in writing .}
\end{itemize}

In such cases all models tend to output the input text as a
prediction.

\section{Results Analysis}\label{sec:error-analysis}

We believe that \textbf{autoencoding} is the most important stage of
the training process. As~\newcite{lample2019multiple} mention, it is a
way to force the model decoder to leverage the style information:
since the noise applied to the input $x$ may corrupt words conveying
the values of the original input style $s$, the decoder has to learn
to use the additional style input in order to perform a proper,
style-infused, reconstruction.

\textbf{Backtranslation} is an easy-to-implement and conceptually
appealing approach: training is straightforward and empirical results
show that it performs well across different datasets and
styles. However, we found that the effectiveness of \bt is not
model-agnostic. We experimented with using a more recent
Transformer~\cite{vaswani2017attention} architecture for the \dae
component and found that the model only manages to do autoencoding,
but almost no style transfer. We hypothesize that this happens when an
encoder's capacity is too high, and is related to the ability of such
models to learn an arbitrary mapping between sequences and associated
latent representation~\cite{shen2019latent}. Prior work for
multi-attribute text style transfer suggests that the encoder is
responsible for encoding the input text into its content
representation~\cite{logeswaran2018content,lample2019multiple}.  In
fact, the interpolated reconstruction loss used in the model
by~\newcite{logeswaran2018content} is based on this assumption.  We
attempted to verify whether the outputs of a Transformer encoder are
used to encourage the content representation of texts rewritten in
different styles to be the same.

\begin{figure}
  \centering
  \includegraphics[width=0.45\textwidth]{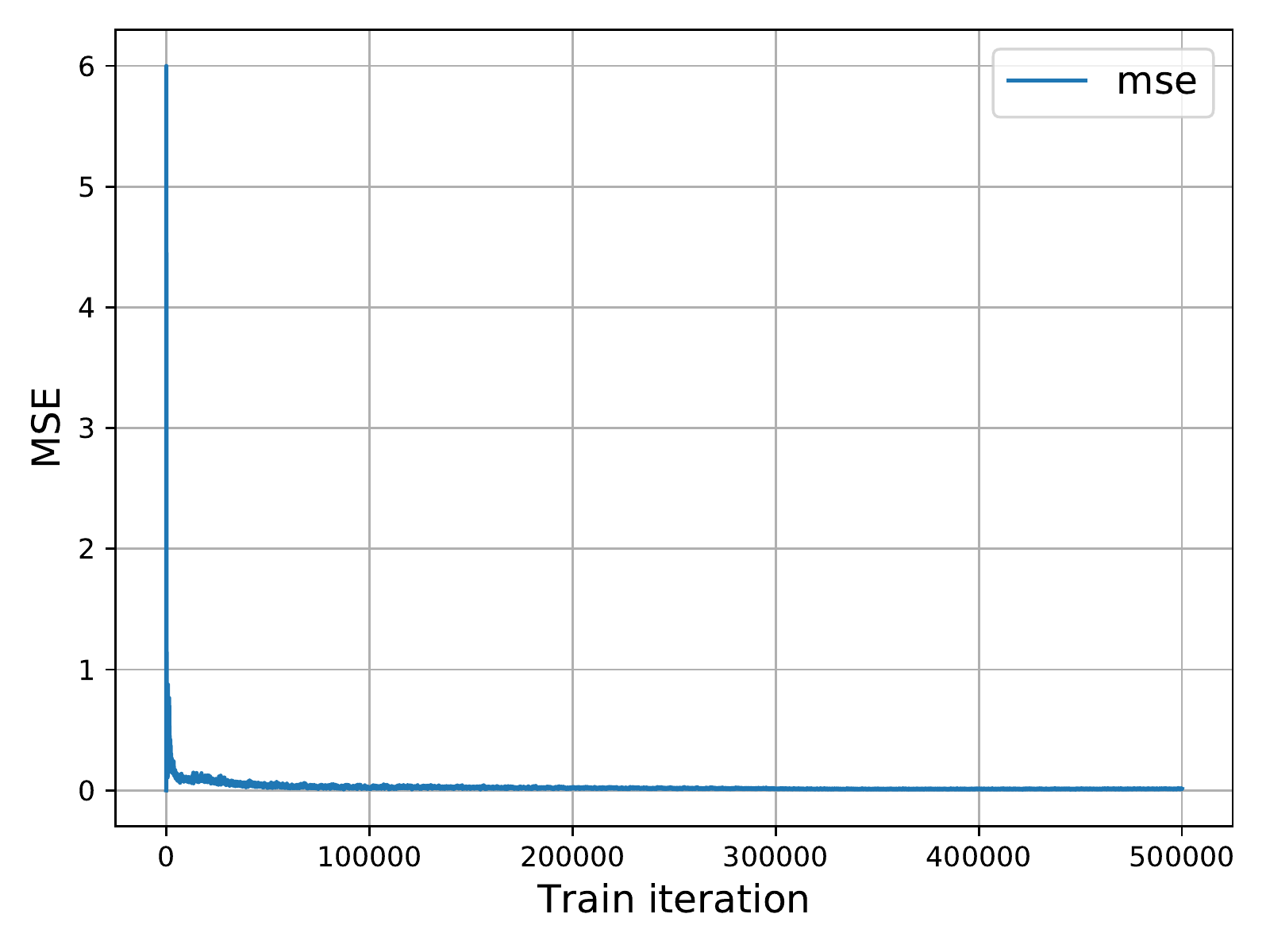}
  \caption{
    Mean squared error between the pooled encoder outputs of the source text and the backtranslated text.
    Development set of the \rt dataset, Transformer-based \dae $+$ \bt model.
  }
  \label{fig:cp-loss}
\end{figure}

During backtranslation, the model generates \xhat and \xstar, which
would be the same text written in different styles, if the model were
perfect.  Assuming that the encoder outputs represent the content, we
can assess how similar the two encoder outputs are. Since the encoder
outputs may have different sequence lengths, we performed a global
pooling over the encoder output vectors, yielding one vector for each
text. Following the single-attribute model
of~\newcite{tikhonov2019style}, we calculate the mean squared error
(MSE) between these two vectors.  The results of this experiment on
the \rt dataset are shown in \Cref{fig:cp-loss}.

The pooled representations become almost the same at the start of
training. Looking back at \Cref{fig:bt}, we can see that it is
possible to "game" the optimization procedure and achieve an optimal
loss value without doing much style transfer. This happens if during
the \dae step the generator model learns to reconstruct the input
without using style information. In this case, \xhat and \xstar in
\Cref{fig:bt} become $x$ and \bt loss becomes 0. Our experiments
suggest that this happens with the Transformer networks and not with
RNN ones. However, the reasons of this phenomenon are not clear.

\textbf{Adversarial training} showed more consistent results in our
experiments, although the training results exhibit more
variation. This is expected --- many researchers reported on the
instability issues with adversarial training, \eg, vanishing
gradients, convergence difficulties,
mode-collapse~\cite{goodfellow2014gan,arjovsky2017principled,roth2017stabilizing}.
Nevertheless, the results are generally lower than the ones we
obtained with \mrt models. A plausible explanation for this could be
the findings of ~\newcite{elazar2018adversarial} who showed that
adversarial discriminators exhibit inferior performance when compared
to external supervision signals.

The \textbf{Minimum Risk Training} method showed both stable training
results and consistent performance gains over the vanilla \bt training
regime. This is a little bit surprising, given that in the neural
machine translation community (where the method is most popular) it is
known to be sensitive to the choice of hyperparameter
values~\cite{shen2016minimum}.

The additional benefit of the \mrt method is that, unlike adversarial
training, one is safe-guarded against the optimization instability
issues: the model is first pretrained with a maximum-likelihood
estimation criterion at the beginning and the worst-case scenario is
staying at the same performance levels. Finally, adversarial
approaches are limited by their use of loss functions that must be
differentiable with respect to the model parameters. \mrt, on the
other hand, can incorporate arbitrary metrics on any level of output
granularity. The biggest weakness of the method is training time ---
getting good parameter estimates depends on the number of samples in
the pool of candidates which are used for approximating the full
search space. As this pool grows, the training time also increases.

\section{Discussion}\label{sec:discussion}

The approaches we examined perform on par, with a slight preference
towards the \mrt method. However, more experiments are needed to
confirm our findings, \eg to understand the strange behavior of the
\trans model trained with \bt. We did not perform additional
experiments comparing the performance of \mrt and \adv models, when
other generator networks (like \trans) are employed. We also did not
experiment with hyperparameter values due to time and computation
constraints, but this is needed in order to account for the randomness
in model training.

Apart from additional experiments explaining the limitations of
backtranslation, we consider the data quality and evaluation protocols
to be two prominent directions that need to be improved.

We found three big issues about the employed datasets. Firstly, with
the exception of the data provided by~\cite{li2018delete}, all other
datasets have multiple versions, which makes model comparison
hard. Secondly, the datasets are centered around style dimensions that
often conflate the \emph{content} and \emph{style} parts. For example,
the multi-attribute Amazon dataset has the review category as an
attribute. However, unlike sentiment transfer, it is not possible to
change the category class of a review without changing its
content. Lastly, some stylistic properties are problematic to model,
\eg, the gender or age of a reviewer. Apart from ethical concerns, we
also found these attributes to be very hard to capture, even by
humans. This means that human evaluation of the models trained on such
data would be problematic.

Evaluation protocols for style transfer models should be improved as
well.  Current metric-based evaluation is flawed for various
reasons. First, the usage of some metrics is questionable. For
example, \bleu is used for measuring content preservation, but it
penalizes differences between input and output texts, even when they
are intended (you cannot change style without changing
content). Second, the reported scores in different works vary even for
the same models. For example, the scores in~\cite{lample2019multiple}
are different from those originally reported in~\cite{li2018delete},
even though model outputs are the same. This most likely happens due
to the differences between the options for training classifers or
computing metric scores (\eg, smoothing method for \bleu). Finally, it
is still not clear what the expected output of a style transfer model
should look like. There is no doubt that a certain trade-off between
content preservation and style transfer intensity is inevitable, but
having some common definition of what constitutes a good model is
definitely needed.

\section{Conclusion}

In this work we empirically compared three most popular approaches to
providing supervision signals in the absence of parallel data for the
task of style transfer. We successfully applied \mrt optimization
techniques to style transfer and showed that it offers the best
performance gains, while staying stable throughout the training. We
revealed a model-specific limitation of the backtranslation method,
which inhibits training style transfer models. We also evaluated a
popular adversarial training approach and found that, although it is
able to improve upon vanilla backtranslation, it does not offer an
advantage over the \mrt alternative.

\section*{Acknowledgments}
This work was supported by the German Federal Ministry of Education
and Research (BMBF) as part of the Software Campus program under the
promotional reference {01IS17050}. The first author of the paper is
supported by the FAZIT Foundation scholarship.

We thank Jessica Ficler for providing us with the \rt data, Raj Dabre
and Munu Sairamesh for the insightful discussions, and our colleagues
Christopher Klamm, Leonardo Ribeiro and G{\"o}zde G{\"u}l {\c{S}}ahin
who provided suggestions that greatly assisted our research.

\bibliography{references}
\bibliographystyle{acl_natbib}

\appendix

\subfile{supp}

\end{document}

%% file: supp.tex
\section{Supplemental Material}
\label{sec:supplemental}

\subsection{Dataset Preparation}\label{sec:supp-dataset-preparation}

\textbf{\syelp} and \textbf{\samazon} datasets are publicly
available.\footnote{\url{https://github.com/lijuncen/Sentiment-and-Style-Transfer}}
The preprocessing steps for \fyelp and \rt datasets are described
below.

\textbf{\fyelp} was prepared using publicly released code.\footnote{\url{https://github.com/facebookresearch/MultipleAttributeTextRewriting}}.
However, due to the computational constraints, we additionally filtered out texts that are longer than 50 tokens. Consequently, this makes our results incomparable to those reported in~\cite{lample2019multiple}. We tried using the same cut-off limit of 100 tokens as in the original paper, but model training became prohibitively expensive.

The raw \textbf{\rt} dataset was shared with us by
\newcite{ficler2017controlling}. We discarded empty reviews, reviews
having only non-alphabetic characters, meta-reviews, and reviews in
languages other than English (the review was considered to be in
English only if at least 70\% of tokens in the text were identified to
be English).

Using available meta-data, we added professionality annotations. We
further followed the instructions of \newcite{ficler2017controlling}
to annotate reviews with their sentiment. As for the gender
annotations, we retrieved them from user names and user ids: we
replaced ids by the actual reviewer names (obtained from the
RottenTomatoes website), and followed the instructions
in~\cite{lample2019multiple} to map the reviewer names to genders
using lists of male/female names.

During training and evaluation all texts were lower-cased.

\subsection{Training Details}\label{sec:supp-training-details}
All models were implemented using PyTorch~\cite{paszke2019pt} and
PyTorch-Lightning\footnote{\url{https://github.com/PytorchLightning/pytorch-lightning}}
frameworks. Our models use the following hyperparameters:
\begin{itemize}
  \item embedding dimension: 512
  \item RNN hidden dimension: 512
  \item encoder pooling kernel size: 5
  \item encoder pooling window size: 5
  \item word shuffle probability: 3
  \item intensity of word shuffling (parameter $k$): 3
\end{itemize}

The models were trained using Adam optimizer~\cite{kingma2014adam} with the following hyperparameters:
\begin{itemize}
  \item lr: 0.0001
  \item betas: (0.5, 0.999)
  \item weight decay: 0
\end{itemize}

The models were trained on a cluster of eight NVIDIA Tesla V100 GPU (32G) for 30 epochs, with a dropout rate of 0.1, gradient norm was clipped to 5.0.
\samazon and \syelp models were trained with a batch size of 400; \rt and \fyelp models used a smaller batch size of 200 due to computational limitations.

We did not restrict the vocabulary size of the models, with an exception of the \fyelp model --- there we followed~\cite{lample2019multiple} and limited the vocabulary to 60k BPE merge operations.

\subsection{Evaluation}\label{sec:supp-evaluation}

All model outputs and references were lower-cased and tokenized by space before evaluation.
Specific details about metrics used are given below:

\textbf{\bleu}, \textbf{\sbleu}. We used the NLTK~\cite{bird-2006-nltk} package to compute \bleu scores. No smoothing was applied.

\textbf{Accuracy}, \textbf{EMD}. We trained fasttext\footnote{\url{https://fasttext.cc/}} classifiers to compute both the accuracy and probability distribution for \emd. We computed the latter using the code from \newcite{mir2019evaluating}. The same codebase was also used to extract style-specific lexicons. 

\textbf{Perplexity}. We used a publicly available KenLM
toolkit\footnote{\url{https://kheafield.com/code/kenlm/}} to train a
5-gram language model with Kneser-Ney smoothing. Perplexities were
computed on the sentence level and averaged over the predicted texts.

\textbf{WMS}. We used the code from \newcite{mir2019evaluating} to compute \wmd scores~\cite{pele2008,pele2009}, but normalised it in the following way:
\begin{equation}
  \label{eq:2}
  \wms(d_1,d_2) = \frac{1}{1+\wmd(d_1,d_2)}  
\end{equation}

Here, $\wmd(d_1, d_2)$ denotes Word Mover's distance between two documents. The reason why we compute the inverse of \wmd is to make it easier for the reader to compare the models: the higher the score, the better the model (similar to the other metrics).
The metric is computed between Word2Vec~\cite{mikolov2013efficient} representations. We used the Gensim Python package~\cite{rehurek_lrec} and trained Word2Vec vectors from scratch on the train portions of the datasets.

\textbf{Excluded metrics}. We excluded some of the metrics that \newcite{mir2019evaluating} originally used in their study. These metrics are:
\begin{itemize}
  \item masked versions of \sbleu and \wms;
  \item adversarial classifiers for measuring naturalness.
\end{itemize}

The former were excluded, because the authors showed that masked versions of the metrics hihgly correlate with unmasked ones. The latter metric was excluded, since the details about training the classifiers were not described in the respective work.